\pdfoutput=1 

\PassOptionsToPackage{dvipsnames,svgnames,table}{xcolor}

\documentclass[sigconf, screen]{acmart}

\setcopyright{none}
\DeclareRobustCommand{\circled}[1]{%
  \tikz[baseline=(char.base)]{
    \node[shape=circle,fill=black,inner sep=0pt,minimum size=1em] (char)
    {\color{white}\normalfont\sffamily\scriptsize #1};}}

\DeclareRobustCommand{\circledna}[1]{%
  \tikz[baseline=(char.base)]{
    \node[shape=circle,draw=black,fill=white,inner sep=0pt,minimum size=1em] (char)
    {\color{black}\normalfont\sffamily\scriptsize #1};}}

\usepackage{url}
\usepackage{amsmath,amsfonts}
\usepackage{graphicx}
\usepackage{textcomp}
\usepackage{subfig}
\usepackage{xcolor} 
\usepackage{multirow}
\usepackage{lipsum}
\usepackage{caption}
\usepackage{wrapfig}
\usepackage{makecell}
\usepackage{xspace}
\usepackage{pifont}

\usepackage[linecolor=DarkGreen,backgroundcolor=DarkGreen!25,bordercolor=DarkGreen]{todonotes}
\usepackage{lipsum}
\usepackage{caption}
\usepackage{booktabs}
\usepackage{lipsum}
\usepackage{adjustbox}
\usepackage{bm}
\usepackage{soul}
\usepackage{listings}
\usepackage{xcolor}
\usepackage[noend, ruled, vlined, linesnumbered, noline, boxed]{algorithm2e}

\newcommand{\tinyscript}{\fontsize{8pt}{9pt}\selectfont}



\definecolor{mdHeader}{RGB}{0, 51, 102}    
\definecolor{mdInline}{RGB}{255, 127, 0}     
\definecolor{mdBlock}{RGB}{128, 128, 128}  
\definecolor{mdLink}{RGB}{0, 102, 204}     

\lstdefinelanguage{Markdown}{
    sensitive=true,
    alsoletter={\#},
    morekeywords={\#,\#\#,\#\#\#,\#\#\#\#,\#\#\#\#\#,\#\#\#\#\#\#},
    keywordstyle=\color{mdHeader}\bfseries,
    morekeywords=[2]{-,*,+},
    keywordstyle=[2]\color{mdHeader}\bfseries,
    moredelim=[s][\color{mdInline}\ttfamily]{`}{`},
    moredelim=[s][\color{mdLink}]{[}{]},
    moredelim=[s][\color{mdBlock}\itshape]{(}{)},
}

\newcommand{\thetitle}{Can AI Review Improve Paper Drafting?\\An Empirical Study on 20 Computer Architecture Submissions}

\begin{document}

\pagenumbering{arabic}

\title{\thetitle}


\author{Di Wu}
\email{di.wu@ucf.edu}
\orcid{0000-0001-9775-8026}
\affiliation{%
  \institution{\mbox{University of Central Florida}}
  \department{Department of ECE}
  \city{Orlando}
  \state{FL}
  \country{USA}
}



\begin{abstract}
Research is advancing faster than ever with artificial intelligence (AI); and so are the corresponding research papers.
The exploding volume of AI-generated papers have put a strain to peer review, leading to the usage of AI-generated review, potentially wide yet sneaky.
However, relevant ethical concerns about confidentiality, quality, and fairness are raised and no consensus has been reached in the broad research community.
We expect the debate to continue for a while, but in the meantime, we ask an alternative, practical question: \textit{can AI review improve paper drafting?}

We study 20 computer architecture papers, with varying levels of submission lineage, to expose how well AI review aligns with human review, quantified by a set of metrics we define.
To conduct the case study, we build a web UI-integrated tool, \emph{AI-Paper-Review}, that generates structured AI review of a draft paper, available at \url{https://github.com/unarylab/ai-paper-review}.
This tool selects several AI reviewers from a diverse pool of AI reviewers and clusters and ranks their comments based on commonality and importance of review comments.
It also allows to align AI comments with human comments to facilitate metric-based validation.
The case study shows that AI review can cover a significant fraction of human-raised issues, but also raises issues missing in human review.

This paper is not intended to encourage using AI for peer review at the current stage, but to study that (1) how AI review can improve paper drafting and (2) the potential and limitation of AI-based peer review.
The release of the tool and the case study data is intended to instigate future research on this topic.
Misuse for peer review would violate the ethics policies from major academic venues.
\end{abstract}

\keywords{artificial intelligence, paper review, paper drafting}

\maketitle
\section{Introduction}
\label{sec:introduction}

\paragraph{Dilema of AI for Peer Review}
AI, mainly large language models (LLMs) at the current stage, is transforming research by unlocking faster discovery in various disciplines~\cite{van2023ai}. 
The result is overwhelmed papers published, causing overloaded peer review.
Given the strain on peer review, we conjecture the wide yet sneaky usage of AI-generated review during peer review, which potentially violates the ethics policies of major academic venues and raises ethical concerns about the confidentiality, quality, and fairness.
Researchers might find it struggling to balance the workload and ethical considerations, both related to AI for research.
Though authors of this paper have not used AI for any peer review so far, we admit it extremely time-consuming to service on the program committee in major computer architecture venues (e.g., ASPLOS, HPCA, ISCA, and MICRO).
Each commitment means reviewing about 15 11-page papers, with each paper review costing about one day.
We have witnessed exploding paper submissions in all computer architecture venues, averaging 30\%-50\% increase compared to the last year.
Similar problems exist in other research communities, perhaps even more severe, especially AI and machine learning.

\paragraph{Alternative Usage of AI Review}
The debate on using AI for peer review is pervasive in many communities and expected to continue for a while.
In the meantime, we ask an alternative, practical question: \textit{can AI review improve paper drafting}, as an ethical application of AI to enhance the research workflow?
More specifically, can it help authors to polish their drafts before submission, and how well does it align with human review?
Answering this question can provide insights into AI-assisted research pipeline: how to use AI review for paper drafting, which is a more ethical and practical usage of AI review at the current stage, and also shed light on the potential and limitation of AI review.

To the best of our knowledge, this is the first empirical study on using AI review for paper drafting, and we hope it can inspire more research on this topic and related topics.

\paragraph{Our Approach}
To answer the question, we conduct a case study on 20 computer architecture papers to expose how well AI review aligns with human review, quantified by a set of metrics we define.
To facilitate this study, we build a web UI-integrated tool, \emph{AI-Paper-Review}, that generates structured AI review of a draft paper.
This tool consists of three major components:
\begin{itemize}
    \item An AI review database that contains a pool of AI reviewers, randomly generated by AI to ensure comprehensiveness and diversity.
    The pool is built from a collection of 200 reviewers, covering 10 sub-domains.
    Each sub-domain contains 20 personas, each grounded in a reviewing preference.
    This database is swappable, either extended to include more sub-domains, or replaced by a customized database for another domain.
    \item A review pipeline that feeds on input paper draft and generate structured review comments per reviewer.
    It first leverage the keywords from the paper and select a few most relevant AI reviewers.
    Then these AI reviewers produce their structured review comments in parallel and independently.
    These comments are then clustered and ranked based on their commonality and importance, allowing authors to prioritize revisions.
    \item A validation pipline that takes in both human and AI review, and aligns AI comments with human comments to validate the effectiveness of AI review.
    The validation pipeline estimates the similarity between each pair of human review comments and AI review comments. 
    It produces qualification metrics such as hits, misses, and false alarms of AI review comments.
\end{itemize}

During the case study, we select 20 computer architecture papers with varying levels of submission lineage and topic diversity to cover a wide range of paper quality and domain.
Though the tool supports multiple AI providers and models, we use Claude Agent SDK for evaluation, and generate AI review with models with different capabilities, while keeping the validation model to the most powerful one, to maximize the quality of AI review and minimize the alignement noise.
The case study shows that AI review can cover a significant fraction of human-raised issues, but also raises issues that human reviewers do not bring up.

This work is unique in terms of its focus on a single discipline and tracking lineage of submissions, both offering more practical insights into the potential and limitation of AI for paper drafting and peer review, unlike prior works that analyze broadly~\cite{liang2024gpt4review}.

\paragraph{Contribution}
The contributions of this paper are listed below.
\begin{itemize}
  \item We systematically study the question of how AI review can improve paper drafting and the potential and limitation of AI-based peer review.
  \item We build a web UI-integrated tool, \emph{AI-Paper-Review}, that generates structured AI review of a draft paper and allows to align AI comments with human comments to validate the quality of AI review.
  \item We conduct a case study on 20 computer architecture papers with varying levels of submission lineage and topic diversity to expose how well AI review aligns with human review, quantified by a set of metrics.
  \item We discuss the implications of the case study and future work on this topic, focusing on the potential and limitation of AI review for paper drafting and peer review.
\end{itemize}

The remainder of the paper are organized as follows.
Section~\ref{sec:background} reviews the usage of AI to enhance research workflow.
Then Section~\ref{sec:architecture} describes the developed AI-Paper-Review tool.
Next, Section~\ref{sec:setup} and~\ref{sec:evaluation} details the evaluation setup and results.
Finally, Section~\ref{sec:discussion} and~\ref{sec:conclusion} discusses and concludes the paper.

\section{Background and Related Work}
\label{sec:background}

\subsection{Dilema of Peer Review System}

\paragraph{Randomness and Bias}
Many authors express concerns about the randomness and bias in peer review, which can lead to unfair and inconsistent outcomes for authors and reviewers~\cite{stelmakh2021prior, rogers2020can}.
Sankaralingam~\cite{10.1145/3686260} categorizes the perceived malfunction of peer review by the reviewer style, and outlines the ``confused'', ``rational'', ``lazy'' and ``Traumatized'' reviewers.
Confused reviewers are not familiar with the topic; rational reviewers could have a conflict of interest with the intention of kill a paper; lazy reviewers do not put enough effort to review a paper; traumatized reviewers want to authors to suffer the same fate, often painful, as themselves before acceptance.
The mixture of different styles of reviewers leads to significant randomness and bias in peer review.

\paragraph{Exploding Submissions}

The existing peer review system is further under crisis, due to the conflict between the huge volume of paper submissions and the relative scarcity of expert reviewers.
Nowadays, it is not uncommon to see a drastic increase in paper submissions compared to the preceding year, across different research communities (e.g., $>2\times$ more submission in ICML 2026 over 2025~\cite{akbashev2026x}).
One of the reason is AI-accelerated paper writing on the authors' side, which is however low quality~\cite{drake2026ai, phys2026ai}.
In major computer architecture venues, authors' perception about the quality of submitted papers before and after the abuse of AI is: \textit{(1) before AI, low-quality submissions are hard to understand, despite of significant efforts; (2) after AI, low-quality submissions are bad obviously in terms of lacking depth.}

Given the randomness and bias of peer review and the explosion of paper submissions, reviewers also turn to AI-generated review for peer review~\cite{drake2026ai, phys2026ai, 10.1145/3686260, sankaralingam2025architectural}.
Sankaralingam~\cite{sankaralingam2025impactmarket} underlines the forming factor as the conflict between ``(1) the rapid dissemination of all sound research and (2) scarce credentialing for prestige and career advancement'', and proposes better incentives for reviewers to address the issue.
Different research communities have distinct policies for AI-assisted peer review.
For example, in the computer architecture community, ISCA 2026 explicitly prohibits the usage of AI for peer review, while ASPLOS 2026 allows it with required disclosure.
We expect this dilemma to continue for a while, due to no consensus reached in the broad research communities.

\subsection{AI for Paper Drafting and Review}
Several lines of work explore AI as reviewers or review assistants.
Checco et al.~\cite{checco2021ai} study about 3 thousands of papers to reveal whether AI can predict the review score of submissions and the potential biases of AI review.
Robertson~\cite{robertson2023gpt4} corroborates the efficacy of AI review at small scale by studying NeurIPS submissions, and observes that GPT-4 can provide useful but moderate feedback with large variation.
Liang et al.~\cite{liang2024gpt4review} analyze the review agreement between GPT-4 and human reviewers for both Nature family journals and ICLR machine learning conference at scale (about 5 thousands of papers in total), and find substantive overlap on AI feedback for weaker papers.
However, the definition of weaker papers is purely based on rejection, while not counting the improvement over time.
Goyal et al.~\cite{goyal2026scholarpeer} build ScholarPeer to synthesize the field’s trajectory, identify omitted state-of-the-art research, and validate claim against other research.

Despite the growing attention in AI drafting and review, there is still a lack of systematic study on how AI review can improve paper drafting precisely.
Our study takes into account the submission lineage of papers to reveal how AI benefits research quality.

\subsection{Persona Prompting}
Persona prompting, instructing AI to ``act as an X'', is well studied.
Shanahan et al.~\cite{shanahan2023role} frame the behavior of an AI agent as a superposition over multiple role-play characters (``simulacra''), with the system prompt acting as a selection mechanism that biases the agent toward a particular character.
Park et al.~\cite{park2023generative} scale up the simulation to multiple agents and build a memory system that can retrieve information, reflect importance, and plan next steps, demonstrating that persona-based agents produce coherent behaviors.
Understanding the usefulness of persona prompting, Zheng et al.~\cite{zheng2024helpful} study the impact of multiple personas on the result quality.
They conclude that simply adding a single persona does not necessarily improve the quality, but adding more personas could be beneficial, while the quality of personas matters more than the quantity.
Olea et al.~\cite{olea2024evaluating} further study the impact of LLM auto-generated personas in open question-answering settings and find that they improve the result quality obviously, if carefully tailored for the question.
De Araujo et al.~\cite{de2025principled} propose principles for designing high-quality personas, such as increasing the expertise of personas, minimizing irrelevant persona details and domain specialization.

With all prior insights, our tool builds a database of diverse AI review personas~\cite{salminen2025using} and facilitate high-quality review by avoiding persona misleading the model~\cite{kim2025persona}.
Similar works are Sankaralingam's three-persona AI review system~\cite{sankaralingam2025reviewer}, including a guardian, a synthesizer, and an innovator to explicitly justify the technical soundness, related work, and future impact.
Our work models these aspects using (1) a large pool of personas, and (2) balanced personas, rather than each agent focusing on a single aspect.

\subsection{AI for Improving Research Workflow}
The exponential growth of scientific output prompts a wave of AI-assisted research.
Zhou et al.~\cite{zhou2025hypothesis}, Chen et al.~\cite{chen2025ai4research}, and Eger et al.~\cite{eger2025transforming} independently taxonomize AI assistance across the full workflow, covering literature search, hypothesis generation, experimentation, writing, and peer review, which demonstrate strong capabilities in automated research workflows~\cite{xu2025deepresearch}.
On the writing side, systems such as SurveyForge~\cite{yan2025surveyforge}, IterSurvey~\cite{cui2025itersurvey}, and AutoSurvey2~\cite{wu2025autosurvey2} automate literature survey generation.
Austin~\cite{austin2025workflow,austin2025heilmeier} documents practitioner patterns for integrating AI agents into daily research tasks.
These works broadly study AI capabilities across the workflow, whereas our work focuses uniquely on whether AI review improves paper quality across the submission lineage.

\subsection{Piloting AI Review Experiments}
STOC 2026~\cite{stoc2026experimental, jayaram2026retrospective} and ICML 2026~\cite{icml2026pat} polits AI review based on Google's Paper Assistant Tool (PAT).
These experiments report 92.1\% of participants find AI feedback helpful and 87.3\% report that AI feedback improves the paper clarity, and 35.4\% of theory-paper authors find gaps requiring more than an hour to fix.
AAAI 2026 also conducts similar experiments and concludes that the majority of participating authors find AI review not only useful, but actually more preferred than human reviews~\cite{aaai_ai_peer_review, biswas2026ai}.

\section{Proposed AI-Paper-Review System}
\label{sec:architecture}

\subsection{Design Principles}
\label{subsec:principles}

Four principles make \emph{AI-Paper-Review} unique.

\emph{Diversity through personas.}
We sample multiple specialized reviewers from a large pool (AI review database), to offer heterogeneous feedback~\cite{zheng2024helpful}.
Each reviewer is comprehensive, rather than a dedicated commentator from a single angle~\cite{goyal2026scholarpeer, sankaralingam2025reviewer, sankaralingam2025architectural}.
This is mimicking the way a program committee review a paper with reviewers of different perspectives.

\emph{Review independence.}
Reviewers review a paper in parallel and in isolation, as human reviewers form opinions before the discussion period.
Only after all review comments are ready, we will analyze the review by clustering and ranking their comments, ultimately offering prioritized comments for authors to fix.

\emph{Validation against ground truth.}
All AI review comments are aligned to the real human review for the same paper, and across submission lineage, letting us answer the title's question with objective quantification rather than subjective human perception in existing experiments~\cite{stoc2026experimental, jayaram2026retrospective, aaai_ai_peer_review, biswas2026ai}.

\emph{Flexibility.}
Despite targeting computer architecture in this paper, our system is not fixed for a single area.
Users can customize the AI review database to meet their own research interests, as well as the number of reviewers based on their needs.

\begin{figure*}[!ht]
  \centering
  \includegraphics{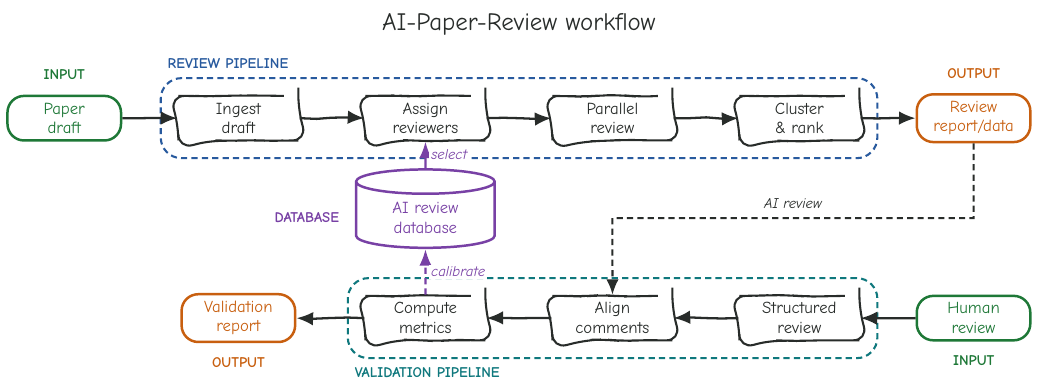}
  \caption{\emph{AI-Paper-Review} workflow.
  The review pipeline ingests a paper draft and the AI review database and produces AI review and the corresponding review report; 
  the validation pipeline aligns the AI review against the human review and outputs a validation report.}
  \label{fig:workflow}
\end{figure*}

\subsection{System Overview}
\label{subsec:system-overview}

Our system has three components as in Figure~\ref{fig:workflow}: a database of reviewers, a review pipeline that turns a draft into ranked comments, and a validation pipeline that aligns those AI comments with human review.
Database calibration is available to suggest how to improve the AI review database.

%
%
%

\subsection{AI Review Database}
\label{subsec:database}

The database is shaped along two dimensions: the subdomain, and the review persona.
Each subdomain is defined by general \texttt{description}, publication \texttt{venues}, and technical \texttt{keywords}.
Next, each persona has a unique review \texttt{focus}, \texttt{style}, \texttt{common\_concerns}, question \texttt{priorities}.
The cross product of subdomains and personas is the total number of reviewers, each carrying one review preference.
Each reviewer is described with a structured header (\texttt{domain}, \texttt{persona}, review \texttt{focus}, \texttt{review style}, and \texttt{keywords}) that is used for review assignment in the review pipeline (Section~\ref{subsec:review-pipeline}), as well as a system prompt that defines \texttt{expertise profile}, \texttt{review lens (persona)}, \texttt{task}, \texttt{output format} and review \texttt{rules}.

For this work on computer architecture, we define 10 subdomains and 20 personas, leading to $10 \times 20 = 200$ reviewers.
Given this huge number of reviewers, it is almost impossible to manually generate them, therefore, we generate them use AI.
Our system automates the generation based on templates, and users just need to specify the types and numbers of subdomains and personas.
Though we use AI generated database, which could be biased, it is possible to calibrate the database using suggestions from the system in the validation pipeline.
However, we are not exploring this in this paper.

\subsection{Review Pipeline}
\label{subsec:review-pipeline}

The review pipeline turns a paper draft into a ranked report in four stages, traced along the top of Figure~\ref{fig:workflow}: it ingests the draft, assigns reviewers from the database, runs them in parallel, then clusters and ranks the resulting comments.
This pipeline is mimicking how a program committee review a paper.

\subsubsection{Paper ingestion}
The pipeline first reads the draft into the text representation every later stage works from.
To extract the title, abstract and body text, a light AI call will be made based on the document head (not full document).

\subsubsection{Reviewer assignment}
The assigner selects the AI reviewers whose expertise best matches the draft from the AI review database.
It embeds the submission (extracted title, abstract, and the first 4000 characters of body text) and each reviewer's keyword profile (keywords plus domain, persona, and focus) with a Sentence-BERT model~\cite{reimers2019sentencebert}, scores every reviewer by cosine similarity, and takes the top $N$ after diversifying personas with $N$ defaulted to 10 and bounded within 1-20.
This diversifying pass ensures AI review gives different perspectives, since reviewers that share a persona use the same review lens and tend to produce similar comments.
The pass walks candidates highest-score-first, skips any persona already chosen so no persona fills two slots.
The pass also caps how many reviewers come from one subdomain, ensuring a few comments from related but not identical subdomains.
The cap allows at most about $0.85\,N$ reviewers from any single subdomain: 6 of 7 at $N{=}7$, or 8 of 10 at $N{=}10$.
The cap is soft: once it is hit, the next in-subdomain candidate is still admitted unless an out-of-subdomain reviewer outscores it by more than $0.15$ in cosine similarity.
Another way for out-of-subdomain reviewers entering the pool is through the persona check: an out-of-subdomain reviewer whose persona is not yet used can outrank some in-subdomain candidate and be admitted.

\subsubsection{Parallel review}
Each selected reviewer is one agent AI call using the system prompt
that returns 5-10 comments in a structured format.
The $N$ calls execute concurrently up to a user-defined limit, default 10.
To lower the cost the call, all reviewers share one system prompt of the paper, followed by the per-reviewer prompt, allowing prefix caching if supported.
Also, an always-on writing-clarity reviewer runs first to warm that cache, and its feedback focuses on non-technical clarity, which will not be used for clustering and ranking.
The AI review call is also resilient, with support for retry upon rate-limit errors and repair upon incorrect output format.
These fixes allow up to 5 times AI re-call, and the low recall rate ensures negligible cost overhead.
The collected per-reviewer comments form the full AI review and feeds the next stage in the review pipeline.

\subsubsection{Clustering and ranking}
This stage takes all raw comments (default $N=10$ reviewers, each producing 5-10 comments) and clusters similar comments to avoid deplications.
To decide what comments belongs together, the pipeline turns each comment's \texttt{summary}, \texttt{description}, and \texttt{keywords} into a meaning-preserving vector with the same Sentence-BERT model used for reviewer assignment, then clusters the comments in order: the first unassigned comment opens a new cluster, and every later comment whose vector is at least $0.55$ cosine-similar to that opener is dropped into the same cluster.
Inside each cluster, the comment with the highest severity (ties broken by the longest description) is kept as the headline, while every other phrasing is preserved in the UI, so that users can see each single comment with similar concerns.

Each cluster $c$ is then scored on how many reviewers agree (commonality) and how serious they think (importance) and ranked, as detailed in Section~\ref{subsec:metrics}.
The score also sets the ranked priority shown to the user.
The ranked clusters are used to generate the review report, ordered from the highest score to lowest, while every original AI comment is kept for the validation pipeline.

\subsection{Validation Pipeline}
\label{subsec:validation}

The validation pipeline measures how well the AI review matches the human review of the same paper, in three stages along the bottom of Figure~\ref{fig:workflow}: it converts the human review into structured comments, aligns them against the AI comments, and computes match metrics.
Unlike prior experiments that ask humans to subjectively rate the AI review's quality~\cite{stoc2026experimental, jayaram2026retrospective, aaai_ai_peer_review, biswas2026ai}, our pipeline gives objective quantification of how close the AI review is to the human review.

\subsubsection{Human-review conversion}
Human reviews come in many forms, while AI reviews follow the structured format from Section~\ref{subsec:review-pipeline}.
Before the two can be compared, the human review is reshaped into the same structure with a single AI call: each concern the reviewer raises becomes one comment block, and the reviewer's wording is copied into that block \emph{verbatim}.
Keeping the wording untouched matters because the next stage decides whether the AI catches a given human concern by measuring how similar the two pieces of text are, paraphrasing here would silently move that score.
On top of the verbatim text, each block is also tagged with a severity (major, moderate, or minor), a category, and the reviewer's overall recommendation, mapped onto the same labels the AI reviewers use, so phrasings like ``weak accept'' or ``5 out of 10'' end up as the same token on both sides.

\subsubsection{Comment alignment}
With both reviews in the same structure, this stage scores how closely the AI and human review matches.
The AI is asked to rate the similarity between every human comment and every AI comment on a $0$--$1$ scale.
For each human comment, its best score then decides a verdict: \emph{same} at $[0.65, 1]$ (essentially a paraphrase), \emph{partial} at $[0.35, 0.65)$ (a related angle on the same comment), and \emph{missed} at $[0, 0.35)$.
Due to potential large numbers of comments on both side, the alignment is chunked into batches, with 5 human comments and all AI comments in one call.
The result is three buckets: hits (human comments the AI catches, including both same and partial matches), misses (human comments the AI misses), and false alarms (AI comments no human brings up).

\subsubsection{Metric computation}
The pipeline counts \emph{hits}, \emph{misses}, and \emph{false alarms}, and condenses them into four metrics, including recall, precision, and their harmonic mean $F_1$, as well as a severity-weighted recall that gives more weight to the misses with higher severity or importance, as detailed in Section~\ref{subsec:metrics}.


Additionally, we can drive an optional calibration step, the dashed path back to the database in Figure~\ref{fig:workflow}, that attributes each miss or false alarm to the persona that should have caught or suppressed it and emits concrete database edits.
We include this capability in the system for completeness but do not exercise it in this study.




\subsection{Implementation}
\label{subsec:implementation}

\emph{AI-Paper-Review} is an open-source Python package, available at \url{https://github.com/unarylab/ai-paper-review}.

\emph{User interface.}
The system can be used two ways.
A Flask web app lets the user upload a PDF, watch the review run live, browse the ranked comments, and optionally upload the corresponding human review to see how well the AI does against it.
The same operations are also command-line tools, including the review pipeline, the validation pipeline, the cross-paper aggregator, the reviewer-database generator, and the web app itself.

\emph{Model layer.}
The same pipeline runs unchanged across many model providers: hosted APIs (Anthropic, OpenAI, Google Gemini, xAI Grok, GitHub Models), SDK-based providers that reuse a local desktop login instead of an API key (Claude Code, GitHub Copilot), and any OpenAI-compatible local endpoint such as Ollama or vLLM.
The review and validation stages pick their model independently from a single YAML config, so the review can run on a strong model while the validation runs on a cheaper one.
PDF ingestion is provider-dependent: APIs receive the original PDF (figures included), while non-APIs might only see text and tables extracted by pypdf / MarkItDown.

\emph{Workflow orchestration.}
The review pipeline is orchestrated as a LangGraph state graph (with a sequential fallback when LangGraph is unavailable), so every stage reads from and writes back to a single shared state record that doubles as the audit trail.

\section{Experimental Setup}
\label{sec:setup}

The case study varies three variables we care, including the paper, the review model, and the reviewer pool size.

\begin{table*}[t]
\centering
\caption{Submission lineage of 20 computer architecture papers.
Each circle means a paper submission.
White/black filled circles are papers that were rejected/accepted at that submission, with submission lineage marked by arrows.
$X$-shot means the number of submissions for a project at year 7, regardless of acceptance or rejection.
Each lineage is one unique project.
}
\label{tab:submission_lineage}
\begin{tabular}{l c c c c c c c c c c c c c c c}
\toprule
\textbf{Submission} & & \textbf{Year 1} & & \textbf{Year 2} & & \textbf{Year 3} & & \textbf{Year 4} & & \textbf{Year 5} & & \textbf{Year 6} & & \textbf{Year 7} & \\
\midrule
\midrule
\textbf{4-shot} & & & & & \circledna{1} & $\rightarrow$ & \circledna{2} & $\rightarrow$ & \circledna{3} & $\rightarrow$ & \circled{4} & & & & \\
\textbf{3-shot} & & & & & & & & & & & \circledna{5} & $\rightarrow$ & \circledna{6} & $\rightarrow$ & \circled{7} \\
\textbf{2-shot} & \circledna{8} & $\rightarrow$ & \circled{9} & & \circledna{10} & $\rightarrow$ & \circled{11} & & & & & & \circledna{12} & $\rightarrow$ & \circledna{13} \\
\textbf{1-shot}  & & & & & & \circledna{14}, \circled{15} & & & & \circled{16}, \circled{17} & & \circledna{18} & & \circled{19}, \circledna{20} & \\
\bottomrule
\end{tabular}
\end{table*}

\begin{table}[t]
\centering
\caption{Topic coverage of 20 computer architecture papers.
Each underline means a unique project, and there are 12 unique projects in total.
}
\label{tab:submission_topic}
\begin{tabular}{l c}
\toprule
\textbf{Topic} & \textbf{Submission} \\
\midrule
\midrule
\textbf{A} & \underline{\circledna{1}, \circledna{2}, \circledna{3}, \circled{4}}, \underline{\circledna{5}, \circledna{6}, \circled{7}}  \\
\textbf{B} & \underline{\circledna{8}, \circled{9}}, \underline{\circledna{10}, \circled{11}}, \underline{\circled{15}} \\
\textbf{C} & \underline{\circledna{12}, \circledna{13}}, \underline{\circled{16}} \\
\textbf{D} & \underline{\circledna{14}}, \underline{\circled{17}} \\
\textbf{E} & \underline{\circledna{18}}, \underline{\circled{19}} \\
\textbf{F} & \underline{\circledna{20}} \\
\bottomrule
\end{tabular}
\end{table}

\subsection{Studied Submissions}

We study 20 real submissions, drawn from \textbf{the authors' own submission history} across all four major computer architecture venues (ASPLOS, HPCA, ISCA, MICRO) and several years.
Using the authors' own papers allows to obtain both original human reviews and its submission lineage, neither of which we could otherwise obtain.
Larger samples would be more representative, but beyond the reach of authors.
The samples exhibits varying submission lineage, unlike prior experiments focusing on submissions to single conferences~\cite{stoc2026experimental, jayaram2026retrospective, icml2026pat, aaai_ai_peer_review, biswas2026ai}.
Table~\ref{tab:submission_lineage} traces the submission lineage, where circled numbers are paper IDs and arrows link successive submissions of one project; the 20 papers collapse into 12 projects, including one 4-shot, one 3-shot, three 2-shot, and seven 1-shot.
Table~\ref{tab:submission_topic} shows the topic coverage, spanning 6 topics.
Paper titles, project topics, and venues are withheld to preserve confidentiality.

\subsection{AI Models}
\label{subsec:models}

We use different models for review and validation to avoid bias.
Review is generated with three model-capability tiers (Opus~4.7, Sonnet~4.6, and Haiku~4.5), so we can measure how model strength affects what the review quality.
The AI reviewers never see the human review to avoid data leak.
Validation is fixed to the strongest model, Opus~4.7, to suppress the validation noise,regardless of the review model.

All calls go through the Claude Agent SDK~\cite{claude_agent_sdk} with the output token limit removed, avoiding output truncation.
This single backend also ensures platform consistency is review and validation.
Review and validation were collected between 2026-05-01 and 2026-05-11, so the whole study sits inside an eleven-day window during which the serving infrastructure for the models may have changed.
Two facts bound the resulting noise: no model version changed over the window~\cite{Anthropic2026Postmortem}, and the window is short relative to the timescale on which served-model quality drifts~\cite{vela2022temporal}.
We therefore treat model behavior as stable across the collection period~\cite{IBM2025ModelCollapse}.

\subsection{AI Review Database}

We build a computer-architecture database, shipped with the tool (Section~\ref{subsec:database}), which includes 200 reviewers, $10$ sub-domains $\times$ $20$ personas.
The database is frozen for the whole study so that review quality reflects the paper and the model.
The selected reviewer pool size $N$ is one experimental knob we sweep: $N\in\{4,7,10\}$, with $N{=}10$ as the default the rest of the study uses unless stated otherwise.

\subsection{Human Review}

The human review is considered as the ground truth we validate AI review against, and we use only human review the authors received on their own papers.
We recover the verbatim human review for 19 of the 20 papers from the submission portals.
The exception is paper~\circled{11}: its portal had closed, so we reconstruct its review from our rebuttal note, which addresses every raised point during the successful revision towards a final acceptance.
The reviewer identities are never disclosed to us under the double-blind policy.
This use of human review stays within venue ethics policies on two counts: it covers only the authors' own submissions, and it is used solely for validation and never released to the public (though risking being uploaded to the AI provider).

\subsection{Assumptions}

The review and validation is grounded on four assumptions.

\begin{itemize}
      \item \textbf{The persona pool is representative.} We assume the selected personas mimicks the realistic reviewer pool; missing personas could systematically bias the review expertise.
      \item \textbf{Importance is categorical.} We bin severity into major, moderate, and minor and weight comments by these bins, assuming identical binning criteria across reviewers.
      \item \textbf{Matching collapses to hit or miss.} The validator assigns each human comment \emph{same}, \emph{partial}, or \emph{missed}, with \emph{same} and \emph{partial} being hits and \emph{missed} being misses.
      \item \textbf{Human review is ground truth.} We treat the human review as the gold standard, even though it is itself subjective and varies across reviewers.
\end{itemize}

\subsection{Evaluation Metrics}
\label{subsec:metrics}

We report two families of metrics: \emph{review-generation} metrics describing the AI review, and \emph{validation-alignment} metrics for measuring how well AI review matches human review.

\paragraph{Review-generation metrics.}
Three scalar scores are reported.
\begin{itemize}
\item 
\emph{Reviewer selection score} is the cosine similarity between the paper embedding and each reviewer's keyword-profile embedding (both produced by Sentence-BERT~\cite{reimers2019sentencebert}), and is what the selector uses to pick the top-$N$ persona-diversified reviewers from the database.
\item
\emph{Comment clustering score} is the cosine similarity between two AI comments, computed by the same Sentence-BERT model over each comment's \texttt{summary $+$ description $+$ keywords}; two comments are placed in the same cluster when this score reaches $0.55$.
\item 
\emph{Cluster ranking score} then ranks each cluster by consensus and seriousness,
$$
\text{score} = n_{\text{distinct}} \cdot
\bigl(0.5\,\overline{w}_{\text{sev}} + 0.5\,w^{\max}_{\text{sev}}\bigr),
$$
where $n_{\text{distinct}}$ is the number of distinct personas that contribute a comment to the cluster, and $\overline{w}_{\text{sev}}$ and $w^{\max}_{\text{sev}}$ are the mean and maximum severity weight in the cluster, with severity weights major $=3$, moderate $=2$, minor $=1$.
\end{itemize}

\paragraph{Validation-alignment metrics.}
For each paper, the alignment step sorts the comments into three groups: a human comment is a \emph{hit} if at least one AI comment matches it (verdict \emph{same} or \emph{partial}) and a \emph{miss} otherwise, and an AI comment is a \emph{false alarm} if no human comment matches it.
Let $n_{\text{human}}$ and $n_{\text{AI}}$ be the total numbers of human and AI comments, and let $n_{\text{hit}}$, $n_{\text{miss}}=n_{\text{human}}-n_{\text{hit}}$, and $n_{\text{false}}$ be the resulting counts; we report four metrics from them.

\begin{itemize}
      \item \textbf{Recall} $= n_{\text{hit}}/n_{\text{human}}$: the share of human concerns the AI matched, what the author would have seen before submission. Higher is better.
      \item \textbf{Severity-weighted recall (SWR)}: recall after weighting each human concern by its severity, so missing a major concern hurts more than missing a minor one,
      \[
      \text{SWR} = \frac{\sum_{h\in\text{hits}} w(h)}{\sum_{h\in\text{human}} w(h)},
      \]
      with weights $w(\cdot)=3$ (major), $2$ (moderate), $1$ (minor); the denominator runs over all human comments (hits plus misses), so a high SWR alongside a lower recall means the AI is catching the important issues even if it misses a few unimportant ones.
      Higher is better.
      \item \textbf{Precision} $= (n_{\text{AI}}-n_{\text{false}})/n_{\text{AI}}$: the share of AI comments that corroborate some human concern. The numerator counts \emph{AI} comments that match, and several AI comments may match the same human one, so it needs not equal $n_{\text{hit}}$; a low precision means the AI raises concerns no human reviewer does. Higher is better.
      \item \textbf{F1} $= 2\,\mathrm{P}\,\mathrm{R}/(\mathrm{P}+\mathrm{R})$: the harmonic mean of precision~$\mathrm{P}$ and recall~$\mathrm{R}$, a single summary of both. Higher is better.
\end{itemize}

\section{Evaluation}
\label{sec:evaluation}

We organize the evaluation in three parts.
The first part prunes the design space, settling on Opus~4.7 as the review model (Section~\ref{subsec:model-impact}) and $N{=}10$ reviewers (Section~\ref{subsec:poolsize}).
The second part walks the review pipeline at the locked (Opus~4.7, $N{=}10$) configuration, testing its three generation scores (Section~\ref{subsec:metrics}) in computation order: reviewer assignment (Section~\ref{subsec:alignment}), comment overlap (Section~\ref{subsec:similarity}), and cluster ranking (Section~\ref{subsec:ranking}).
The third part validates the review against human ground truth, from broad recovery to the human decision: recovery (Section~\ref{subsec:coverage}), comment fidelity (Section~\ref{subsec:fidelity}), coverage of the most severe concerns (Section~\ref{subsec:critical}), misses and additions (Section~\ref{subsec:failure}), agreement with human acceptance recommendation (Section~\ref{subsec:verdict}), and shift with draft maturity (Section~\ref{subsec:readiness}).

\subsection{Model tier and reviewer count}
\subsubsection{How does the model tier change the review quality?}
\label{subsec:model-impact}

\begin{figure}[t]
  \centering
  \includegraphics[width=\columnwidth]{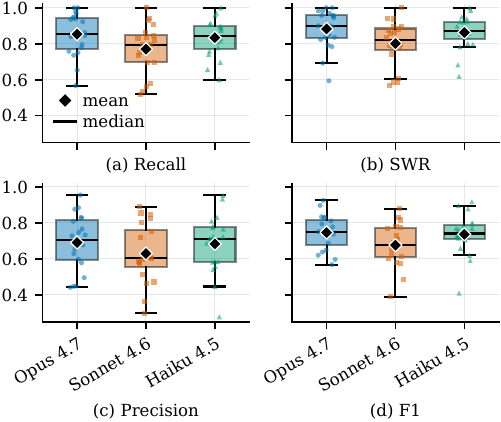}
  \caption{Review quality by model tier at the pool size $N{=}10$.
  Each panel is one metric over the three review models.
  Box spans the interquartile range (IQR; the 25th-75th percentile) across the 20 papers with the median as the inner line, dots $=$ the papers (marker shape per model tier), black diamond $=$ mean, whiskers reach $1.5\times$IQR beyond the quartiles.
  }
  \label{fig:model_tier}
\end{figure}

Figure~\ref{fig:model_tier} compares the three review models at the pool size $N{=}10$.
Mean recall is $0.85$ for Opus~4.7, $0.83$ for Haiku~4.5, and $0.77$ for Sonnet~4.6.
SWR follows recall's ordering and widens the Opus lead ($0.88$ for Opus~4.7, $0.86$ for Haiku~4.5, and $0.80$ for Sonnet~4.6), so the strongest model is also the one that best catches the concerns that matter most.
The smaller Haiku surpasses Sonnet in review quality, but the advantage of Opus~4.7 is clear enough.
Precision is lower and flatter than recall, holding in the $0.63$-$0.69$ band across the three tiers: a stronger review model raises true matches and false alarms together, with Opus~4.7 and Haiku~4.5 essentially tied.
F1 lands between them and preserves the same tier ordering, Opus~4.7 $>$ Haiku~4.5 $>$ Sonnet~4.6.
Opus~4.7 leads on almost every metric, and we use it for the rest of the evaluation.
The superiority of Opus~4.7 in review is also why we use it for validation.

\subsubsection{How many reviewers are enough?}
\label{subsec:poolsize}

\begin{figure}[t]
  \centering
  \includegraphics[width=\columnwidth]{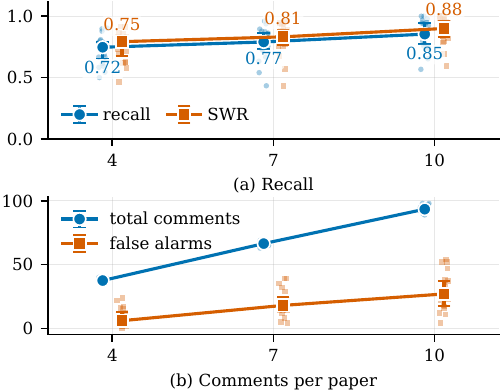}
  \caption{Effect of the reviewer pool size $N$ with Opus~4.7 for both and validation.
  \textbf{(a)} Recall and SWR versus $N$.
  Each dot is for one paper (recall as circles, SWR as squares), markers are the median, error bars span the IQR, and the number at each bar is the mean.
  \textbf{(b)} Total comments and false-alarm comments per paper versus $N$, with the same point/median/IQR encoding.}
  \label{fig:poolsize}
\end{figure}

It is expected that more AI reviewers will raise more comments and cover more human review comments.
Figure~\ref{fig:poolsize} shows the experiments on how adding more reviewers impacts review quality, by sweeping $N \in \{4,7,10\}$ on the Opus~4.7.
Mean recall climbs $0.72 \to 0.77 \to 0.85$ across the three pool sizes, with SWR moving in lockstep from $0.75$ to $0.88$; neither plateaus inside the tested range.
The improvement of recall from $N{=}7$ to $N{=}10$ ($+0.08$) is larger than from $N{=}4$ to $N{=}7$ ($+0.05$), still bending up rather than flattening at the top of the swept range.
Comment volume per paper also increases, with the mean number of AI comments per paper growing from 38 to 94 and false alarms from 9 to 29.
This indicates that a larger pool benefits coverage at the price of more non-issues.
Based on the results, we fix $N{=}10$ as the default for the rest of the evaluation.

\subsection{Review pipeline}

The review pipeline runs in three stages, and we evaluate each stage against its own score from Section~\ref{subsec:metrics}.

\subsubsection{Are the assigned reviewers aligned with the paper?}
\label{subsec:alignment}

\begin{figure}[t]
  \centering
  \includegraphics[width=\columnwidth]{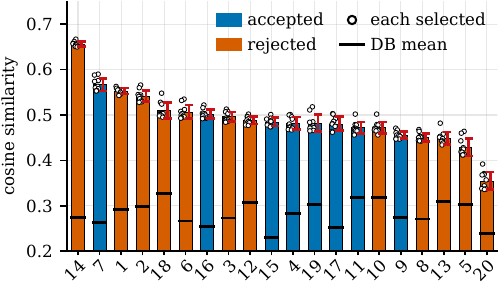}
  \caption{Reviewer selection score on Opus~4.7 at $N{=}10$, per paper.
  Bar height $=$ mean cosine similarity of the ten selected reviewers; white dots on the left half $=$ the ten individual scores; red error bar on the right half $=\,\pm 1$ standard deviation across the ten; black tick $=$ database-wide mean across all 200 reviewers (random-draw baseline).
  Bars are colored by submission outcome.}
  \label{fig:alignment}
\end{figure}

The selector earns its place at the front of the pipeline.
Across the 20 papers the assigned reviewers score a mean cosine similarity of $0.49$, far above the $0.28$ database-wide mean a random draw from the 200-reviewer database would yield, and the pool diversifies along personas while staying topically focused.
Figure~\ref{fig:alignment} reports the per-paper picture: each bar is the mean similarity of the ten selected reviewers, the small white dots on the left half of the bar are those ten individual scores, the red error bar on the right half shows the same mean with a $\pm 1$ standard-deviation spread, and the black tick on each bar is the database-wide mean across all 200 reviewers (random-draw baseline).
The top-1 reviewer in the database is the topmost dot on every paper, because persona-diversification does not skip the top-1 on any of the 20 papers.

Three patterns hold across the sample.
First, top-1 and pool mean track within 0.02 of one another on every paper (means 0.51 and 0.49 across the 20), therefore, the persona-diversifying pass sacrifices negligible alignment to gain persona diversity for better review quality.
Second, the improvement over the random baseline is large and consistent: a mean of $+0.21$ with a minimum of $+0.12$ on \circledna{20} and a maximum of $+0.38$ on \circledna{14}, with the variation induced by the AI review database subdomain coverage.
Third, the number of assigned sub-domains is small: 13 of the 20 papers' pools sit in a single sub-domain, 6 span two, and only \circledna{20} reaches three.

Across the 20 papers the Pearson correlation between top-1 similarity and recall is $-0.07$ (and $-0.04$ against SWR).
Once the reviewer assignment is done, more aggressive alignment does not capture more human comments.
Furthermore, alignment is likewise unrelated to submission outcome: accepted and rejected papers draw equally well-matched pools.

\textbf{Takeaway:} Reviewer assignment is well aligned with paper topics, but the alignment score does not predict recall or submission outcome.

\subsubsection{How similar are the generated AI comments?}
\label{subsec:similarity}

\begin{figure}[t]
  \centering
  \includegraphics[width=\columnwidth]{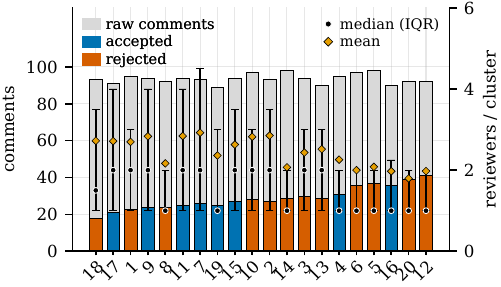}
  \caption{Comment clustering on Opus~4.7 at $N{=}10$, per paper, sorted by compression ratio.
  Left axis for comments: gray $=$ raw AI comments, colored bar $=$ clustered comments (colored by submission outcome); the gap between them is the redundancy the comment-clustering score removes.
  Right axis for reviewers per cluster: black circle $=$ median reviewers with the vertical bar its IQR, orange diamond $=$ mean.}
  \label{fig:similarity}
\end{figure}

The generated comments from multiple reviewers are often redundant, with some being one reviewer restating a point and some being cross-reviewer consensus.
Scoring every pair of AI comments by cosine similarity and merging any pair at or above $0.55$ into a shared cluster (Section~\ref{subsec:metrics}) collapses an average of $94$ comments per paper (89-98) into an average of $29$ clusters (18-41), an average compression of $3.4\times$ (2.2-5.2$\times$), as shown in Figure~\ref{fig:similarity}.
The redundancy is largely genuine consensus rather than one reviewer repeating itself: a cluster draws on $2.4$ distinct reviewers on average (1.8-2.9).
The distribution is skewed: the median cluster holds a single reviewer, but clusters with more than one average $3.8$ reviewers.
Accepted and rejected papers compress by about the same ratio (mean $3.5\times$ vs.\ $3.3\times$, median $3.6\times$ vs.\ $3.3\times$), and their clusters are shared by a similar number of distinct reviewers (mean $2.6$ vs.\ $2.3$, median $2.7$ vs.\ $2.3$).
These gaps are not statistically significant across the 20 papers.
One exception is cluster count: accepted papers trend toward slightly fewer clusters (mean $26.9$ vs.\ $30.1$, median $25.5$ vs.\ $29.0$), though this gap too is not significant (Pearson correlation $r{=}{-}0.26$).

\begin{figure}[t]
  \centering
  \includegraphics[width=\columnwidth]{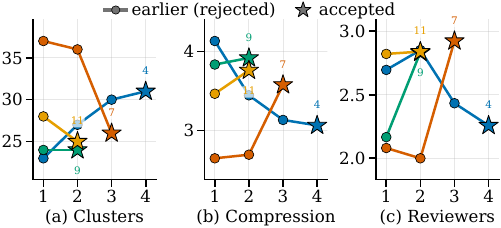}
  \caption{Comment clustering along submission lineage on Opus~4.7 at $N{=}10$.
  Each colored trajectory is one project whose lineage ended in acceptance, drawn left-to-right by submission round; the accepted submission (star) carries its paper ID.
  \textbf{(a)} clusters per paper, \textbf{(b)} compression ratio, and \textbf{(c)} mean distinct reviewers per cluster.
  Star $=$ the accepted submission, filled circle $=$ an earlier rejected submission.}
  \label{fig:lineage-similarity}
\end{figure}

Figure~\ref{fig:lineage-similarity} traces four projects whose lineage ended in acceptance from their first submission to their last.
The three shorter lineages (stronger submissions) compress more and pool more distinct reviewers per cluster when near acceptance, but the long-submission lineage (weaker submission; \circledna{1}, \circledna{2}, \circledna{3}, \circled{4}) reverses all three, accumulating clusters ($23\to31$), losing compression ($4.1\times\to3.1\times$), and drawing on fewer reviewers per cluster ($2.7\to2.3$) as it matures.

\textbf{Takeaway:} The comment-clustering score collapses raw comments into fewer consensus-based clusters.
While accepted papers trend toward fewer clusters with weak statistical significance, neither the compression ratio nor the reviewers per cluster predicts submission outcome.
None of these three shifts consistently across a project's submission lineage, but three shorter lineages (stronger submissions) do show the expected pattern of more compression and more reviewer consensus as they near acceptance.

\subsubsection{Does the cluster ranking surface human concerns first?}
\label{subsec:ranking}

\begin{figure}[t]
  \centering
  \includegraphics[width=\columnwidth]{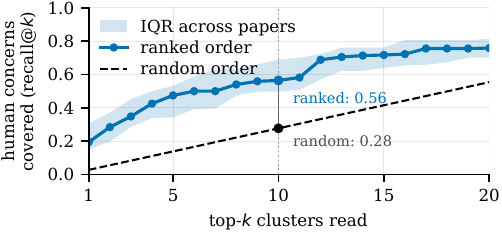}
  \caption{Ranking quality on Opus~4.7 at $N{=}10$.
  recall@$k$ is the share of the caught human concerns that have appeared after reading the top $k$ ranked clusters, as the median across the 20 papers (band $=$ IQR) against a random reading order.
  }
  \label{fig:ranking}
\end{figure}

Our system ranks comments worth acting on near the top for authors to prioritize.
After clustering, the pipeline orders the clusters by their importance and commonality (Section~\ref{subsec:metrics}).
Figure~\ref{fig:ranking} shows how well that ranking surfaces the human concerns the AI catches, measured by recall@$k$: the share of the caught human concerns that have appeared after reading the top $k$ ranked clusters.
Reading in the ranked order surfaces concerns far faster than a random order: the top 10 ranked clusters already cover $0.56$ of the caught concerns, against just $0.28$ for ten clusters read at random, about twice as many for the same reading effort.
Half of the caught human concerns appear within the top six clusters and $70\%$ within thirteen.

\textbf{Takeaway:} The cluster ranking score surfaces the caught human concerns faster than a random order, so authors can skim only the top dozen clusters and still recover most human concerns.

\subsection{Validation Pipeline}
\subsubsection{How much human review does AI recover?}
\label{subsec:coverage}

\begin{figure}[t]
  \centering
  \includegraphics[width=\columnwidth]{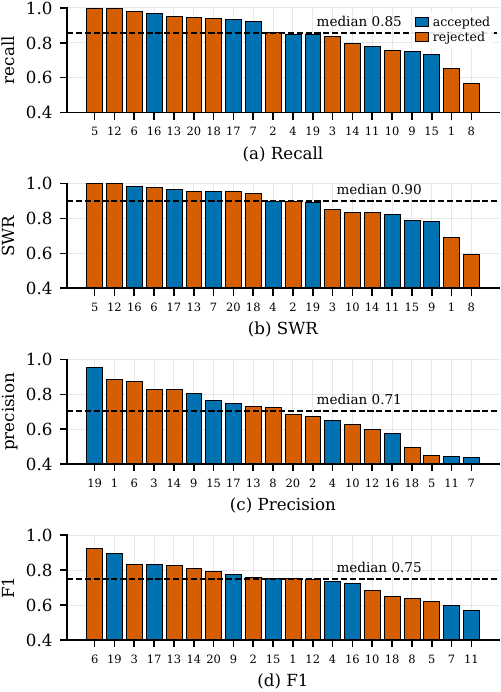}
  \caption{Coverage of human review on Opus~4.7 at $N{=}10$.
  \textbf{(a)}-\textbf{(d)} sorted per-paper recall, SWR, precision, and F1, with the median marked and bars colored by submission outcome.}
  \label{fig:coverage}
\end{figure}

Figure~\ref{fig:coverage} reports all validation metrics.
Recall and SWR form a coverage axis.
Across all papers median recall is $0.85$ (mean $0.85$, range $0.57$-$1.00$) and median SWR is $0.90$ (mean $0.88$, range $0.59$-$1.00$).
SWR exceeds plain recall on every paper (Pearson correlation $r{=}0.99$), since AI catches more severe comments better.
Precision and F1 form a false-alarm axis, where F1 tracks precision well due to high recall.
Recall and precision tend to trade off across papers, i.e., papers where the tool recovers more human comments also tend to carry a higher false-alarm rate.
For example, high recall papers (5 and 12) fall in the bottom third on precision, and three of the four papers below $0.5$ precision (5, 7, and 18) sit above the median on recall.
Finally, neither axis distinguishes accepted papers from rejected ones: all four metrics are statistically indistinguishable by outcome.

\textbf{Takeaway:} AI review recovers most human comments, especially the severe ones, but it also raises many non-issues; and neither the coverage nor the false-alarm rate tracks submission outcome.

\subsubsection{How faithful are the recovered comments?}
\label{subsec:fidelity}

We further look into how well the AI comments align with the human comments caught.
At validation the aligner scores every AI-human comment pair on a $0$-$1$ similarity scale and records, for each caught concern, the similarity of its best-matching AI comment ($\mathrm{primary\_sim}$) together with the two comments' severity and category.
Over the $576$ caught comments, the comment-alignment score has a median of $0.65$ (IQR $0.55$-$0.80$), so a matched AI comment is a close paraphrase of the human concern.
Beyond matching the concern, the AI also tracks its severity and rarely softens it.
The matched pair carries the same severity label on $42\%$ of hits, and on the $332$ disagreements the AI rates the concern more severe than the human far more often than less ($233$ versus $99$).
Counting agreements and upgrades together, the AI is at least as severe as the human on $83\%$ of caught concerns.

\textbf{Takeaway:} A recovered concern is a faithful paraphrase of the caught human comment, and the AI almost never underestimates the severity of the caught concerns.

\subsubsection{Does AI review catch the decision-critical concerns?}
\label{subsec:critical}

\begin{figure}[t]
  \centering
  \includegraphics[width=\columnwidth]{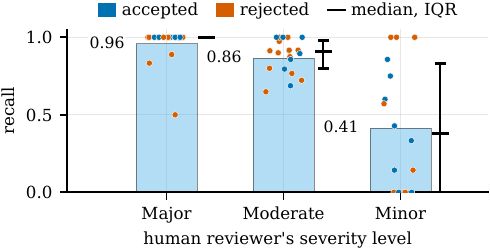}
  \caption{Recall by the human reviewer's own severity level on Opus~4.7 at $N{=}10$.
  Bars are pooled recall across all 20 papers (hits divided by hits${+}$misses within each level); points are per-paper recall, colored by submission outcome; the black whisker to the right of each bar marks the per-paper median and interquartile range (25th-75th percentile).}
  \label{fig:critical}
\end{figure}

We further look into how well the AI review aligns with the severity level of human review, with results shown in Figure~\ref{fig:critical}.
Across the 20 papers, the pooled recall for major severity is $0.96$ ($139$ of $145$), with the per-paper median being $1.00$.
Furthermore, $17$ of the $20$ papers catch all major concerns, and the worst-covered paper recovers half of the major concerns ($0.50$).
The pooled recall then falls to $0.86$ on moderate concerns and $0.41$ on minor ones, with per-paper medians of $0.91$ and $0.38$.
These results, based on the human reviewer's own severity rather than AI labels, suggest that recall climbs with the human's own severity label from minor to major.

\textbf{Takeaway:} AI review focuses on the decision-critical concerns and rarely misses a concern that would sink the paper.
The concerns it does drop are overwhelmingly minor rather than decisive.

\subsubsection{What does AI review miss, and what does it add?}
\label{subsec:failure}

\begin{figure}[t]
  \centering
  \includegraphics[width=\columnwidth]{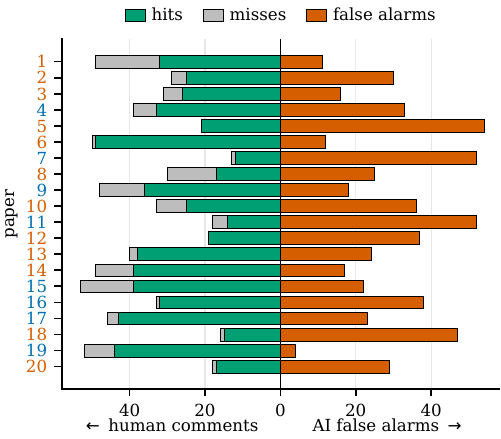}
  \caption{What AI review catches, misses, and adds on Opus~4.7 at $N{=}10$.
  Human comments extend left of zero, split into \emph{hits} (caught by the AI) and \emph{misses} (uncaught); AI comments that matched no human concern (\emph{false alarms}) extend right.
  Splitting the two sides at zero keeps the human-comment total (hits${+}$misses) distinct from the AI additions.
  Paper IDs are colored by submission outcome (blue $=$ accepted, vermillion $=$ rejected).
  }
  \label{fig:failure}
\end{figure}

Figure~\ref{fig:failure} accounts for every comment on each paper.
On all 20 papers the left bar is mostly hits, so the AI misses few human comments; the variation is on the right, where the number of extra AI comments ranges from $4$ to $54$ across papers.
What the AI misses is little and mostly minor.
Of the $111$ missed human comments, only $6$ are major, and the rest are moderate ($64$) or minor ($41$).
Looking into the calibration suggestions, the reason why a comment is missed: in $83\%$ of cases ($92$) no invited reviewer covered that angle, and in only $17\%$ ($19$) an invited reviewer saw the draft but stayed silent.
What the AI adds is plentiful.
It raises $29$ extra comments per paper on average, and each extra comment is not automatically wrong but maybe a real concern never written down.

\textbf{Takeaway:} The AI misses little, and what it misses could be a question of which reviewers we invite rather than weak reasoning; it adds many comments, many of which may be valid concerns the human reviews left unrecorded rather than noise.

\subsubsection{Does AI review track the human verdict?}
\label{subsec:verdict}

\begin{figure}[t]
  \centering
  \includegraphics[width=\columnwidth]{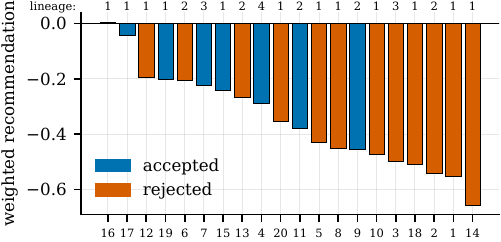}
  \caption{Whether the AI panel's recommendation tracks the human verdict on Opus~4.7 at $N{=}10$.
  Each of the ten pool reviewers emits one recommendation; the score is scaled by the reviewer's selection similarity and averaged per paper, sorted, and colored by outcome.
  The number at the top is the paper's submission index within its project lineage.}
  \label{fig:verdict}
\end{figure}

Each of the ten reviewers emits an overall recommendation, which we map to a signed score: strong accept, accept, and weak accept to ${+}2$, ${+}1.5$, ${+}1$; borderline to $0$; and weak reject, reject, and strong reject to ${-}1$, ${-}1.5$, ${-}2$.
We scale each reviewer's score by the reviewer selection score to mimick reviewer expertise (Section~\ref{subsec:metrics}) and average the scaled scores over the ten reviewers, with results drawn in Figure~\ref{fig:verdict}.
The weighted recommendation is negative on $19$ out of the $20$ papers and only one accepted paper reaches a positive value.
Understanding that the AI review tends to be more severe, it is reasonable that the selected AI reviewers exhibits a higher bar towards acceptance.
Ranked by that recommendation, accepted papers sit well above rejected ones (weighted recommendation $-0.23$ versus $-0.43$), meaning that the AI review reliably separates the better-prepared drafts from the weaker ones.

\textbf{Takeaway:} The AI panel does not predict acceptcance, but it well ranks accepted drafts above rejected ones, so its recommendation is best read as a relative quality signal across drafts.

\subsubsection{How does AI review change as a draft matures?}
\label{subsec:readiness}

\begin{figure}[t]
  \centering
  \includegraphics[width=\columnwidth]{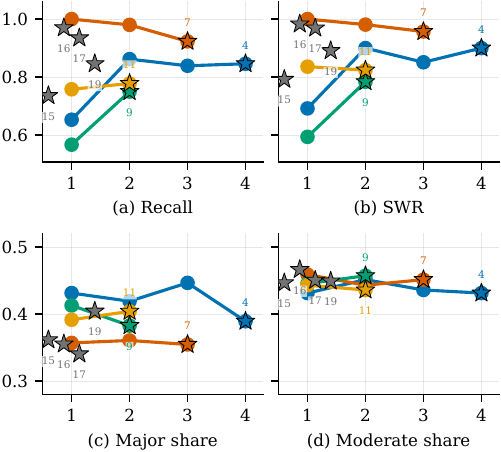}
  \caption{AI-review quality across submission lineage on Opus~4.7 at $N{=}10$.
  \textbf{(a-b)} recall and SWR are the two $0$-$1$ scores and share the y axis.
  \textbf{(c-d)} the major share and the moderate share of the AI comments, sharing the same y axis.
  All versus submission version in x axis; multi-shot projects are colored trajectories from early draft (circle) to the accepted version (star), and single-shot accepted papers are gray stars.}
  \label{fig:lineage}
\end{figure}

We hypothesize that AI review narrows as a draft matures, recovering more human concerns while shifting away from major weaknesses after revisions.
Figure~\ref{fig:lineage} tests this hypothesis.
Across the multi-shot projects (one 4-shot, one 3-shot, two 2-shot; Table~\ref{tab:submission_lineage}) we track recall and SWR over successive submissions, together with how the AI-comment severity shifts.
The maturity signal is real but uneven, concentrating where the first draft has room to improve.
On the two projects that start weakest, the AI catches more human concerns as the draft is revised: recall rises from $0.65$ to $0.85$ over the 4-shot project (\circledna{1}-\circled{4}) and from $0.57$ to $0.75$ over a 2-shot project (\circledna{8}-\circled{9}), with SWR improving alongside it.
The 3-shot project (\circledna{5}-\circled{7}) that is already well captured on its first try has nowhere left to climb, so it stays high or slips a little ($1.00 \to 0.92$).
What the AI flags also shifts on the improving projects: the share of its comments marked as major weaknesses drops ($0.43 \to 0.39$ and $0.41 \to 0.38$), while the share marked moderate barely moves.

\textbf{Takeaway:} As a draft matures, the review coverage rises and the major comments falls, especially on papers with weak starts.

\section{Discussion}
\label{sec:discussion}

\subsection{A Drafting Aid, Not a Reviewer}
\label{subsec:ai-review}

What the tool catches matters far less than who is running it.
When authors run it on their own draft, it does what we expect: it flags problems early, while there is still time to fix them.
When someone runs it on a paper assigned to them for review, the common concerns about secrecy, quality, and fairness all apply, and offloading to local AI review does not solve them.
Another problem is more concerning: if authors can run the tool on their own, \textit{what reviewers bring to the process intellectually}.

\subsection{Beyond Computer Architecture}
\label{subsec:communities}

Computer architecture is a small community, and the overload is far worse in AI and machine learning.
Nothing in this work is tied to computer architecture except the pool of reviewers we use: swap the pool and the tool points at a new field, while the review, alignment, and scoring mechanisms stay the same.
Extending the study to another field then comes down to gathering a new database and a set of papers whose human reviews are known, which is exactly what the open-source release aims at.

\subsection{Recommendation to the Community}
\label{subsec:suggestion}

The computer architecture community could allow author-side AI feedback while keeping the ban on reviewer-side AI, similar to what AI communities are trying~\cite{stoc2026experimental, jayaram2026retrospective}.
Pairing that with a disclosure rule would let the community keep the drafting benefit we measure without giving up the intellectual judgment that peer review exists to provide.
This author-side-only model is already being trialed in parts of the computer architecture community.
Incentive stays unaddressed: until reviewing earns the credit that authorship does, the submission volume that makes AI review tempting will keep climbing~\cite{sankaralingam2025impactmarket}.

\section{Conclusion}
\label{sec:conclusion}

AI is accelerating research to unprecedented levels, causing excessive load in peer review.
Although AI review remains controversial for ethical concerns, it can still help polish paper drafts before submission.
We build a web UI-integrated tool, \emph{AI-Paper-Review}, that generates a structured AI review of a draft and aligns it with human review to validate review quality.
We evaluate it in a case study on 20 computer architecture papers with varying levels of submission lineage and topic diversity.
The case study shows that AI review can cover a significant fraction of human-raised issues, but also raises issues that human reviewers do not bring up.

\begin{acks}
AI is used in the preparation of this manuscript for editing, grammar checking, and knowledge retrieval. 
No passages were copied without full author review, and all substantive ideas, analyses, and conclusions are the product and responsibility of the authors. 
Additionally, AI is utilized for code development and early paper review.
\end{acks}


\bibliographystyle{ACM-Reference-Format}
\bibliography{refs}

\end{document}